\documentclass[acmlarge,screen,nonacm]{acmart}
\AtBeginDocument{%
  }

\setcopyright{acmlicensed}
\copyrightyear{2025} 
\acmYear{2025} 
\setcopyright{acmlicensed}\acmConference[GECCO '25]{Genetic and Evolutionary Computation Conference}{July 14--18, 2025}{Malaga, Spain}
\acmBooktitle{Genetic and Evolutionary Computation Conference (GECCO '25), July 14--18, 2025, Malaga, Spain}
\acmDOI{10.1145/3712256.3726327}
\acmISBN{979-8-4007-1465-8/2025/07}

\usepackage{booktabs, multirow}

\begin{document}

\title{Adaptive Estimation of the Number of Algorithm Runs in Stochastic Optimization}

\author{Tome Eftimov}
\email{tome.eftimov@ijs.si}
\orcid{0000-0001-7330-1902}
\affiliation{%
  \institution{Computer Systems Department \\ Jožef Stefan Institute}
  \city{Ljubljana}
  \country{Slovenia}
}

\author{Peter Korošec}
\email{peter.korosec@ijs.si}
\affiliation{%
  \institution{Computer Systems Department \\ Jožef Stefan Institute}
  \city{Ljubljana}
  \country{Slovenia}
}

\renewcommand{\shortauthors}{Eftimov and Korošec}

\begin{abstract}

Determining the number of algorithm runs is a critical aspect of experimental design, as it directly influences the experiment's duration and the reliability of its outcomes. This paper introduces an empirical approach to estimating the required number of runs per problem instance for accurate estimation of the performance of the continuous single-objective stochastic optimization algorithm. The method leverages probability theory, incorporating a robustness check to identify significant imbalances in the data distribution relative to the mean, and dynamically adjusts the number of runs during execution as an online approach.

The proposed methodology was extensively tested across two algorithm portfolios (104 Differential Evolution configurations and the Nevergrad portfolio) and the COCO benchmark suite, totaling 5,748,000 runs. The results demonstrate 82\%–95\% accuracy in estimations across different algorithms, allowing a reduction of approximately 50\% in the number of runs without compromising optimization outcomes. This online calculation of required runs not only improves benchmarking efficiency, but also contributes to energy reduction, fostering a more environmentally sustainable computing ecosystem.
\end{abstract}



\maketitle

\section{Introduction}

Estimating the performance of the algorithm is a time-consuming process and requires the repetition of numerous algorithm runs on different problem instances. This is especially important for the algorithm's development phase, where we need to establish if the newly developed algorithm is significantly different, hopefully, better than the previous version or some other competing algorithms. This is most commonly achieved through benchmarking~\cite{bartz2020benchmarking}. To derive a valid conclusion about algorithm performance, one needs to carefully select problem instances, establish experimental setups, and assess performance in accordance with benchmarking theory. The selection of the number of algorithm runs is one of the most crucial decisions in experiments setup, which influence the time required to perform the experiment and the validity of the conclusions drawn from it. If the number of experimental runs is insufficient, critical characteristics of the algorithm's performance may not be adequately captured. Conversely, an excessive number of runs could lead to inefficient use of resources, resulting in unnecessary time expenditure without a corresponding increase in insight.~\cite{vevcek2017influence}. So the question that we would like to answer is, is there a way to determine if the number of runs so far performed (i.e., estimated after each run) provides enough information to draw valid conclusions, so no further runs would be needed?

The state-of-the-art performance evaluation is based on reporting descriptive statistics (e.g., mean, median, standard deviation) or comparing performance distributions~\cite{eftimov2017novel}. Therefore, establishing enough information by performing enough algorithm runs is crucial. Given the stochastic nature of the algorithms, where each run may yield inherently different outcomes, it is theoretically optimal to maximize the number of runs to ensure comprehensive and reliable performance evaluation. Looking at journal papers and competitions held at different conferences, this number can be from 15~\cite{bbob} to 51~\cite{liang2013problem}, with common values set at 25 or 30. The critical questions to address are: Are 15 runs sufficient to capture the necessary performance characteristics, or is that already unnecessarily high? Alternatively, is 51 excessive, resulting in an inefficient allocation of time and resources? Ideally, the number of runs should be minimized while still ensuring it is adequate to produce statistically robust and reliable performance evaluations. 

In reality, the number of runs does not need to be the same for all problem instances.
For problem instances where the algorithm demonstrates consistent behavior, a lower number of runs may suffice to obtain reliable performance estimates. Conversely, for problem instances where the algorithm exhibits highly variable or unpredictable behavior, a greater number of runs is necessary to ensure statistically robust evaluations. 
However, this issue is not solely dependent on the characteristics of the problem instance but is also influenced by the algorithm's performance on that specific problem instance. Thus, it is possible for one algorithm to exhibit consistent performance on a given problem instance, requiring fewer runs to achieve reliable evaluation, while another algorithm may display highly variable behavior on the same problem instance, necessitating a greater number of runs to obtain statistically robust results.

\noindent\textbf{Our contribution:} Given the critical importance of selecting an appropriate number of runs for valid and efficient algorithm performance evaluation, this paper proposes an empirical methodology for determining the required number of runs to ensure accurate analysis in continuous single-objective stochastic optimization. The approach uses probability theory, especially the symmetry or balance of the data distribution around its mean.It can be assumed as an online approach that estimates the number of runs during the execution phase of the experiment. This means there is no a priori number of runs set, but the execution of runs is stopped when the information gathered from so far runs is sufficient enough to represent the performance of the algorithm, which, in single-objective problems, is the quality of the best solution. The proposed methodology has been evaluated on a large set of experiments conducted by pairing two different algorithm portfolios (104 Differential Evolution (DE) configurations~\cite{price2006differential} and 11 Nevergrad algorithms~\cite{nevergrad}) by using the 24 problems from the Black-Box Optimization Benchmarking (COCO)~\cite{bbob} in different problem dimensions including 10, 20, and 40. The total amount of runs executed is 5,748,000. The experimental results have shown that the approach led to approximated 82\% - 95\% correct estimations depending on the combination of an algorithm portfolio and a benchmark suite. These results led to a rough estimation of reducing the number of runs conducted in this study and still providing accurate continuous single-objective stochastic optimization analysis. It comes out that approximately 50\% of the overall runs conducted in this study can be omitted. Encouraging the online calculation of the number of runs required for accurate analysis for a specific algorithm and problem instance fosters experiments aimed at reducing energy usage and promoting an environmentally conscious computing ecosystem, also known as green benchmarking.

\noindent\textbf{Outline:} The remaining sections of the paper are structured as follows: Section~\ref{s:RW} outlines the related work; Section~\ref{s:NR} presents the empirical approach to determining the required number of runs for reliable single-objective continuous optimization; Section~\ref{s:experimental_design} provides a detailed explanation of the experimental design, focusing on the algorithm and problem portfolios used in the study; Section~\ref{s:results} presents the evaluation results of the proposed methodology; Section~\ref{s:green} addresses the green benchmarking aspect of the proposed methodology. Lastly, Section~\ref{s:conclusions} concludes the study and suggests directions for future research.

\noindent\textbf{Data and code availability:} The data and the code used in this study are available at \url{https://zenodo.org/records/15099850}.

\section{Related work}
\label{s:RW}
Estimating the number of runs for stochastic optimization algorithms is a challenging task, as it depends on several factors such as the problem size, the algorithm parameters, the stopping criteria, the desired accuracy, and the statistical significance. A common approach is to use a fixed number of runs for each problem and compare the average results of different algorithms. Recently, a study~\cite{vermetten2022analyzing} shows that an insufficient number of runs can impact the effectiveness of automated algorithm configuration methods, leading to the possibility of selecting a suboptimal configuration by chance. The study demonstrates that relying solely on mean performance values, a common practice among configurators, necessitates a substantial number of runs to obtain reliable comparisons among the configurations being considered.

In most cases, the conventional approach recommends using ``standard" values for the number of runs, typically 30 or 50, or even more~\cite{derrac2011practical}. From a statistical perspective, increasing the number of runs enhances the likelihood of detecting significant performance differences between the algorithms being compared~\cite{bartz2007experimental}. Larger sample sizes increase the sensitivity of statistical methods, allowing the detection of even minor differences. However, this heightened sensitivity can lead to the misinterpretation of negligible effects with no practical significance as being statistically significant~\cite{bartz2006new,eftimov2019identifying}. 
It is important to emphasize that statistical analyses conducted with moderately sized or even small samples can yield results that are equally valuable and robust as those derived from very large representative samples~\cite{ng2022unbiggen}, provided the experimental design is rigorous, the assumptions underlying the statistical tests are met, and the sample is appropriately representative of the population of interest.

In~\cite{campelo2019sample}, the authors introduce a methodology for determining the necessary sample sizes (i.e., number of problem instances and number of repeated runs) when designing experiments with specific statistical properties for comparing two methods within a given problem class. The proposed approach enables researchers to specify desired levels of accuracy for estimating mean performance differences on individual problem instances, as well as the desired statistical power for comparing mean performances across the problem class. This approach determines the number of runs ($n_i$, $i=1, 2$) of two algorithms $A_1$ and $A_2$ on a specific problem instance as the problem of identifying the minimum total sample size, $n_1 + n_2$, required for the standard error of the means difference estimator to be below a predetermined accuracy threshold. This problem is a subject of constraint optimization. It is important to highlight that in this approach, the number of runs is always considered within the context of comparing two or multiple algorithms (the number of runs depends on the combination of which algorithms are paired). Furthermore, the authors present a generalization of the approach for comparing multiple algorithms~\cite{campelo2020sample}. However, in our study, our focus is not on comparing algorithms but rather on developing an empirical online approach that can determine when enough data has been collected from running an algorithm on a specific problem instance, neglecting the behavior of other algorithms on the same problem instance.
\section{Estimating the number of runs needed}
\label{s:NR}

We propose an empirical method to determine the number of runs required for reliable single-objective continuous optimization. Our goal is to estimate the number of runs, $n$, necessary to gather sufficient representative data from executing an algorithm on a specific problem instance. This ensures that the $n$ runs, $(x_1, x_2,\dots, x_n)$, provide a robust assessment of the algorithm's performance.

To estimate this, we begin by executing the algorithm $p$ times on a given problem instance, where $p$ may correspond to a predefined threshold (the selection will be explained further). From the resulting $p$ values, we compute their mean and center each value by subtracting this mean. 
This operation is presented below:
\begin{equation}
    Y_i= X_i - \overline{X}.
    \label{eq:first}
\end{equation}
where $\overline{X}$ is the mean value from the $p$ runs. 
From probability theory, it follows that by subtracting the mean of $p$ values from the same distribution 
the resulting variables are not independent Bernoulli variables. The $Y_i$ are not independent because of their shared dependency of $\overline{X}$. 


To determine the required number of runs, we conduct a robustness check, by evaluating the symmetry or balance of the data distribution around its mean. For this reason, we use the skewness of the $Y_i$ variables as a non-parametric diagnostic to assess whether the data distribution exhibits significant imbalance relative to the mean. Our hypothesis is that an algorithm gathering sufficient data from the runs should exhibit a symmetric distribution around the mean. 

The pipeline for estimating the required number of runs is outlined as follows:

\noindent\textbf{(1) Initial Runs:} Execute the algorithm instance on a given problem instance five times, $p=5$, collecting the solution values $(x_1, \dots, x_5)$. 
The decision to use five initial runs is arbitrary, following a practice also adopted in other statistical methods unrelated to our case, such as the $\chi^2$ test of goodness of fit~\cite{cochran1952chi2}. From these solution values, compute the transformations $Y_i$.

\noindent\textbf{(2) Symmetry Check:} Assess the symmetry of the data. 
Calculate the skewness, $\tilde{\mu}_3 = \mathbb{E} \left[ \left( \frac{Y - \mu}{\sigma} \right)^3 \right]$,  of the $Y_i$ values. In an ideal symmetric scenario, the skewness should be close to zero. To accommodate minor deviations, a predefined threshold ($\tau$) around zero is set to evaluate whether the distribution can be considered symmetric ($-\tau \le \tilde{\mu}_3 \le \tau$).
   
\noindent\textbf{(3) Adjust Runs Based on Symmetry:} If the skewness falls outside the predefined threshold, execute an additional run ($p = p + 1$) and repeat all necessary calculations to reassess the distribution's symmetry. If the skewness lies within the predefined threshold, terminate the process. The needed number of runs is $n=p$.

Skewness can be used to assess the symmetry of a distribution; however, in practice, an algorithm instance's performance may be significantly impacted by a few outliers (a few extreme run values not consistent with the majority of runs because of the stochastic nature of the algorithms, that can be either on one or both sides of the distribution) that strongly influence the skewness. To address this, empirical adjustments are necessary, involving the application of techniques to manage outliers, ensuring the effectiveness of the proposed approach.

To apply the proposed approach, an additional preprocessing step is conducted to address the presence of outliers prior to performing the estimation. Given a sample of size $p$, denoted as $(x_1, x_2, \dots, x_p)$, an outlier detection technique is applied before executing the calculation steps and evaluating the symmetry of the distribution. If the technique identifies $m$ outliers, these observations are excluded, resulting in a refined sample of size $p-m$ used for the symmetry check. Notably, the final estimator for the required number of runs, provided the predefined conditions are met, is still based on the original sample size $p$.

We selected three well-established techniques for handling outliers, which are described in detail below:

\noindent \textbf{(1) Interquartile range (IQR) method}~\cite{whaley2005interquartile} -- It is a measure of the spread of data. In general, the value $\mathit{IQR}$ is the difference between the 25th ($q_{0.25}$) and 75th ($q_{0.75}$) percentiles of the data. When it is used for outlier detection, all data values below $q_{0.25} - 1.5 \cdot \mathit{IQR}$ and above $q_{0.75} + 1.5 \cdot \mathit{IQR}$ are considered outliers.

\noindent \textbf{(2) Percentiles-based method}~\cite{diciccio1988review} -- In this method, all data values that are outside the 2.5th and the 97.5th percentiles will be detected as outliers.

\noindent \textbf{(3) Modified z-score}~\cite{kannan2015labeling} -- It is a more robust way to detect outliers than a z-score. The z-score, which provides information on how many standard deviations a value is from the mean value, can be affected by unusually large or small data values. The modified z-score is calculated as $modified-z-score=\frac{0.6745(x_i - \tilde{x})}{\mathit{MAD}}$, where $x_i$ is a single data value, $\tilde{x}$ is the median of the data, and $\mathit{MAD}$ is the median absolute deviation of the data. If the $\mathit{MAD}=0$, then we use the mean of the absolute deviation of the data instead of $\mathit{MAD}$ for division. The values with modified z-scores that are less than -3.5 or greater than 3.5 are detected as outliers.

\section{Experimental design}
\label{s:experimental_design}
Here, we outline the experimental design used to collect data for evaluating the proposed methodology, detailing the benchmark suite and optimization algorithms included in the study.

\noindent\textbf{Problem portfolio:} We utilized the COmparing Continuous Optimizers (COCO)~\cite{hansen2021coco} problem suite, which comprises 24 single-objective optimization problem classes. For each class, an arbitrary number of instances can be generated, with instances representing shifted, scaled, and/or rotated variants of the same problem. From the suite's available problem dimensions, we selected $D \in \{10, 20, 40\}$.

\noindent\textbf{Algorithm portfolio:} For optimization algorithms, the primary evaluation was conducted using the standard Differential Evolution (DE) algorithm~\cite{price2006differential} on the COCO benchmark suite, using 15 problem instances. To generate diverse data reflecting varying algorithm behaviors, 104 DE instances were created by randomly varying hyperparameters: mutation strategy, scaling factor, and crossover probability. Mutation strategies were selected from the following pool: Rand/1/Bin, Rand/1/Exp, Rand/2/Bin, Rand/2/Exp, Best/1/Bin, Best/1/Exp, Best/2/Bin, Best/2/Exp, Best/3/Bin, Rand\\/Rand/Bin, RandToBest/1/Bin, and RandToBest/1/Exp. The scaling factor $F$ and crossover probability $Cr$ were chosen from the interval $(0, 1)$. The population size was set equal to the problem dimension, with stopping criteria defined as one of the following: $D \times 10,000$ function evaluations, $100$ iterations without improvement, or achieving an optimum solution within $10^{-8}$ of the true optimal value.

To mitigate potential bias from focusing on a single family of optimization algorithms, additional experiments were conducted using 11 algorithms from Nevergrad~\cite{nevergrad}. The selected algorithms include Differential Evolution (DE)~\cite{de}, Diagonal CMA~\cite{hansen2001self_adaptation_es}, NaivelsoEMNA~\cite{EDAbook}, NGOpt14, NGOpt38 (two versions of Nevergrad's built-in algorithm selection wizard~\cite{meunier2022black}), OnePlusOne~\cite{beyer2002evolution}, modCMA~\cite{hansen2001self_adaptation_es}, modDE~\cite{vermetten2023modular}, PSO~\cite{pso}, Random Search, and RCobyla~\cite{cobyla}. All algorithms were run with default hyperparameter values, with stopping criteria set to $D \times 2,000$ function evaluations. This stopping criteria differs from our DE experiments because the data was reused from a publicly available study~\cite{kostovska2023gecco}. The data contains the performance of the Nevergrad algorithms using 10 instances per problem for a single dimension, $D = 20$.

\noindent \textbf{Performance data:} Each algorithm configuration was executed 50 times for each problem instance and corresponding dimension. For the DE configurations and the COCO benchmark suite, we utilized 1,080 unique triplets (24 problems $\times$ 15 instances $\times$ 3 problem dimensions). Each triplet was executed 50 times with 104 DE configurations, resulting in a total of 5,616,000 runs. For the Nevergrad experiments, 240 triplets (24 problems $\times$ 10 instances $\times$ 1 problem dimension) were tested 50 times with 11 algorithms, ending up in 132,000 runs. Altogether, this experimental setup produced 5,748,000 algorithm runs, generating optimization results (errors to the global optimum) used to evaluate the proposed methodology.

\noindent \textbf{Evaluation scenarios:} To evaluate the accuracy of estimating the required number of runs, we generated 50 independent solution values ($x_1, x_2, \dots, x_{50}$) for each combination of a triplet (problem, instance, and dimension) and an algorithm instance, forming a ground truth sample. Using this sample, we sequentially analyzed subsets starting with the first five solutions to determine the required number of runs, $n$. The mean difference between the $n$-sized sample and the ground truth sample mean was computed to estimate the sample size effect. To assess precision and confidence in this estimation, we calculated the bootstrap confidence interval (CI) of the mean difference.

\noindent\textbf{Bootstrapping evaluation scenario}:
Since outlier detection techniques are part of our estimation approach, the evaluation begins by applying an outlier detection method to the estimated $n$-sized sample $(x_1, x_2, \dots, x_n)$. Let $m_\textrm{e}$ be the number of detected outliers, resulting in a sample of size $n - m_\textrm{e}$. Similarly, the same outlier detection technique is applied to the 50 ground truth samples, yielding a size of $50 - m_\textrm{t}$ after removing $m_\textrm{t}$ outliers.  Next, $M$ resamples of size 50 are drawn from both the $(n - m_\textrm{e})$ and $(50 - m_\textrm{t})$ samples, and their mean differences are computed. Using these $M$ mean differences, the 2.5th and 97.5th percentiles are calculated to construct a 95\% percentile confidence interval (CI). While the bootstrapped mean difference typically follows a normal distribution by the Central Limit Theorem~\cite{dudley1978central}, skewed distributions may occur. In such cases, the bias-corrected and accelerated bootstrap~\cite{diciccio1988review} is used to account for skewness. By examining whether the CI contains zero, we determine the percentage of cases where the number of runs estimation leads to accurate optimization analysis. For cases where the analysis is inaccurate, the bias-corrected and accelerated bootstrap is further checked to handle skewness and verify if the CI contains zero.

\noindent\textbf{Post-hoc empirical evaluation:} If neither the original confidence interval nor the bias-corrected and accelerated confidence interval includes a zero mean difference, we calculate an empirical measure to assess how often the bootstrapped mean value from the estimated sample falls within a specified percentage of the bootstrapped mean value from the ground truth sample. The following percentage thresholds are used: $0.50\%$, $1.00\%$, $5.00\%$, $10.00\%$, $15.00\%$, and $20.00\%$.

\section{Results}

\label{s:results}

\noindent\textbf{Comprehensive Benchmarking of DE Configurations Across Problem Dimensions:}
Table~\ref{tab:tab_1} shows the percentage of triplets (problem, instance, and dimension) where the proposed method accurately estimates the number of runs. The ``True" column represents percentages based on bootstrapping CIs (original and bias-corrected), while the other six columns reflect post-hoc evaluation within specified error ranges. The results are based on 104 DE configurations tested on the COCO benchmark suite (24 problem classes, 15 instances per class, and three dimensions: $D \in \{10, 20, 40\}$). The evaluation, based on bootstrapping CIs requiring stochastic sampling, was repeated 10 times for each triplet and configuration using different sampling seeds. The percentages for each configuration were averaged across these repetitions, and the final values were averaged over all configurations, as shown in Table~\ref{tab:tab_1}.

\begin{table}[!htp]\centering
\caption{The percentage of triplets where the proposed method accurately estimates the number of runs. The ``True" column shows results from bootstrapping CIs, while the next six columns indicate cases within specified error margins (as labeled).}\label{tab:tab_1}
\scriptsize
\begin{tabular}{lrrrrrrrrr}\toprule
\multicolumn{1}{c}{\shortstack{Skewness \\ threshold}} & 
\multicolumn{1}{c}{\shortstack{Outlier \\ technique}} & 
 true & $\le$0.5\% & $\le$1\% & $\le$5\% & $\le$10\% & $\le$15\% & $\le$20\% \\\midrule
\multirow{3}{*}{0.05} & IQR & 84.13 & 84.17 & 84.22 & 86.12 & 88.73 & 90.85 & 92.66 \\
& Percentile & 79.91 & 79.93 & 79.97 & 82.47 & 85.86 & 88.44 & 90.48 \\
& MAD & 86.16 & 86.20 & 86.24 & 87.66 & 89.77 & 91.48 & 93.00 \\\midrule
\multirow{3}{*}{0.10} & IQR & 77.21 & 77.26 & 77.33 & 79.94 & 83.62 & 86.62 & 89.15 \\
& Percentile & 72.36 & 72.38 & 72.44 & 75.79 & 80.38 & 83.96 & 86.80 \\
& MAD & 79.72 & 79.78 & 79.83 & 81.86 & 84.89 & 87.34 & 89.46 \\\midrule
\multirow{3}{*}{0.15} & IQR & 72.91 & 72.97 & 73.04 & 76.09 & 80.41 & 83.90 & 86.88 \\
& Percentile & 67.86 & 67.89 & 67.97 & 71.81 & 77.11 & 81.30 & 84.57 \\
& MAD & 75.68 & 75.75 & 75.81 & 78.23 & 81.78 & 84.69 & 87.21 \\\midrule
\multirow{3}{*}{0.20} & IQR & 69.50 & 69.57 & 69.64 & 73.01 & 77.74 & 81.64 & 84.95 \\
& Percentile & 64.63 & 64.66 & 64.74 & 68.94 & 74.66 & 79.26 & 82.77 \\
& MAD & 72.37 & 72.44 & 72.52 & 75.21 & 79.15 & 82.45 & 85.26 \\
\bottomrule
\end{tabular}
\end{table}

The approach was evaluated for four skewness thresholds (0.05, 0.10, 0.15, and 0.20) using three outlier detection techniques. Among the 104 DE configurations, the \emph{MAD} technique achieved the highest accuracy (86.16\%), correctly estimating 933 out of 1,080 triplets based on bootstrap CIs at a skewness threshold of 0.05. The \emph{IQR} and \emph{Percentile} techniques yielded similar results with slight reductions in accuracy. For triplets where the correct number of runs was not estimated via bootstrap CIs, post-hoc evaluation was applied. With the \emph{MAD} technique at a 0.05 skewness threshold, cumulative accuracy increased to 87.66\% within a 5\% error of the ground truth and 89.77\% within a 10\% error. The results show that increasing the skewness threshold, which allows greater flexibility and deviations from distribution symmetry, leads to a decrease in estimation accuracy as anticipated. However, accuracy remains high ($\ge$ 74.00\%) when considering solutions within a 10\% error margin of the ground truth sample mean.

\begin{figure}[]
    \centering
    \includegraphics[scale=0.42]{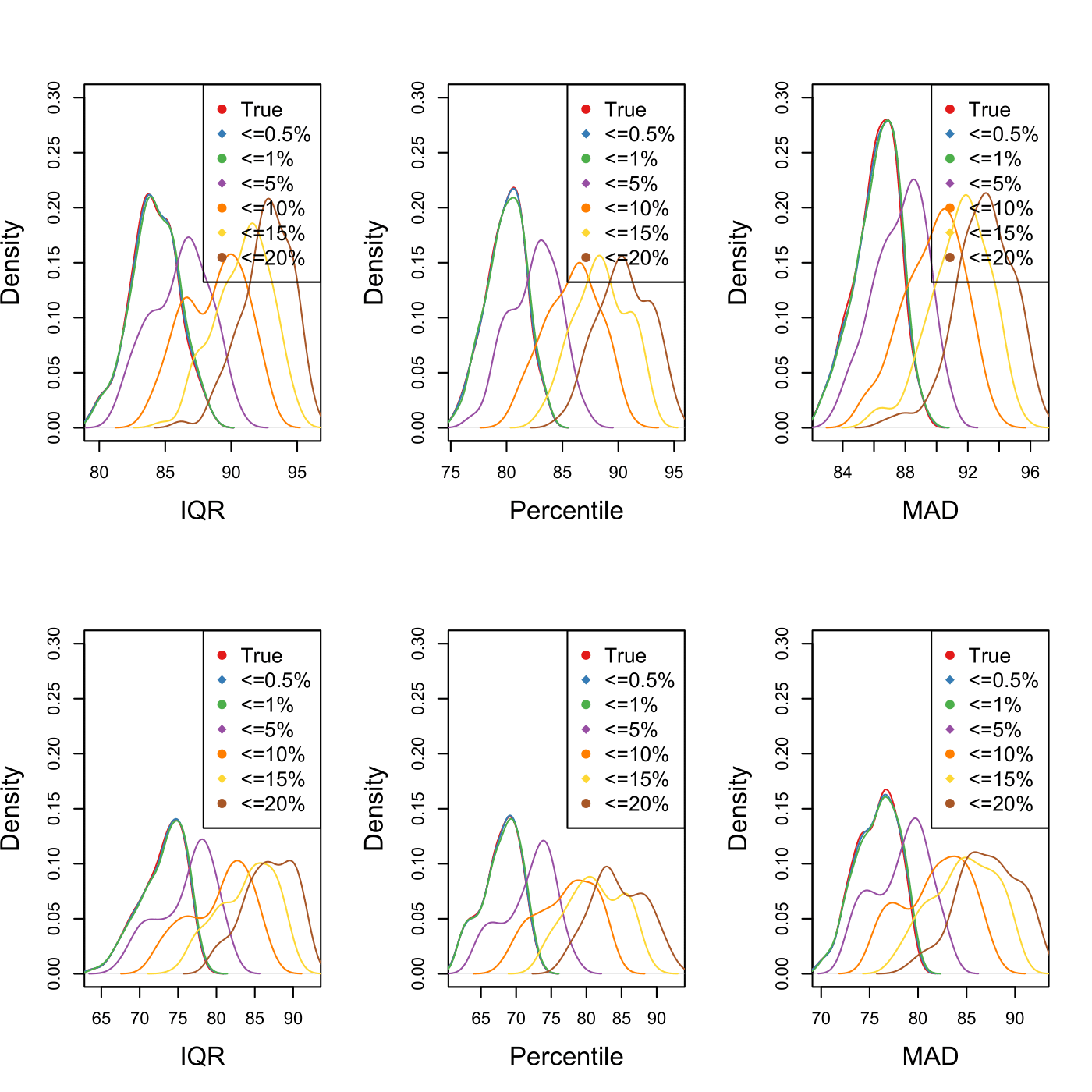}
    \caption{
The distribution of correctly estimated run percentages across all 104 configurations is presented for a skewness threshold of 0.05 (top row) and 0.15 (bottom row). The columns correspond to different outlier detection techniques (\emph{IQR}, \emph{Percentile}, and \emph{MAD}).}
    \label{fig:all_alg_all_dim}
\end{figure}

Figure~\ref{fig:all_alg_all_dim} provides additional insights into the percentages of correctly estimated runs across all 104 DE configurations for skewness thresholds of 0.05 (top row) and 0.15 (bottom row). Most configurations show percentages of correctly estimated triplets near the average reported in Table~\ref{tab:tab_1}. For triplets with incorrect run estimations, post-hoc evaluation reveals that as the error range increases (e.g., $\leq 5\%$, $\leq 10\%$, $\leq 20\%$), the percentage of triplets shifts to higher values across all configurations. The patterns are similar also for the other skewness thresholds and are not presented due to the page limit. Similar patterns are observed for other skewness thresholds but are omitted here due to space constraints.

\begin{figure*}[t]
    \centering
    \includegraphics[scale=0.5]{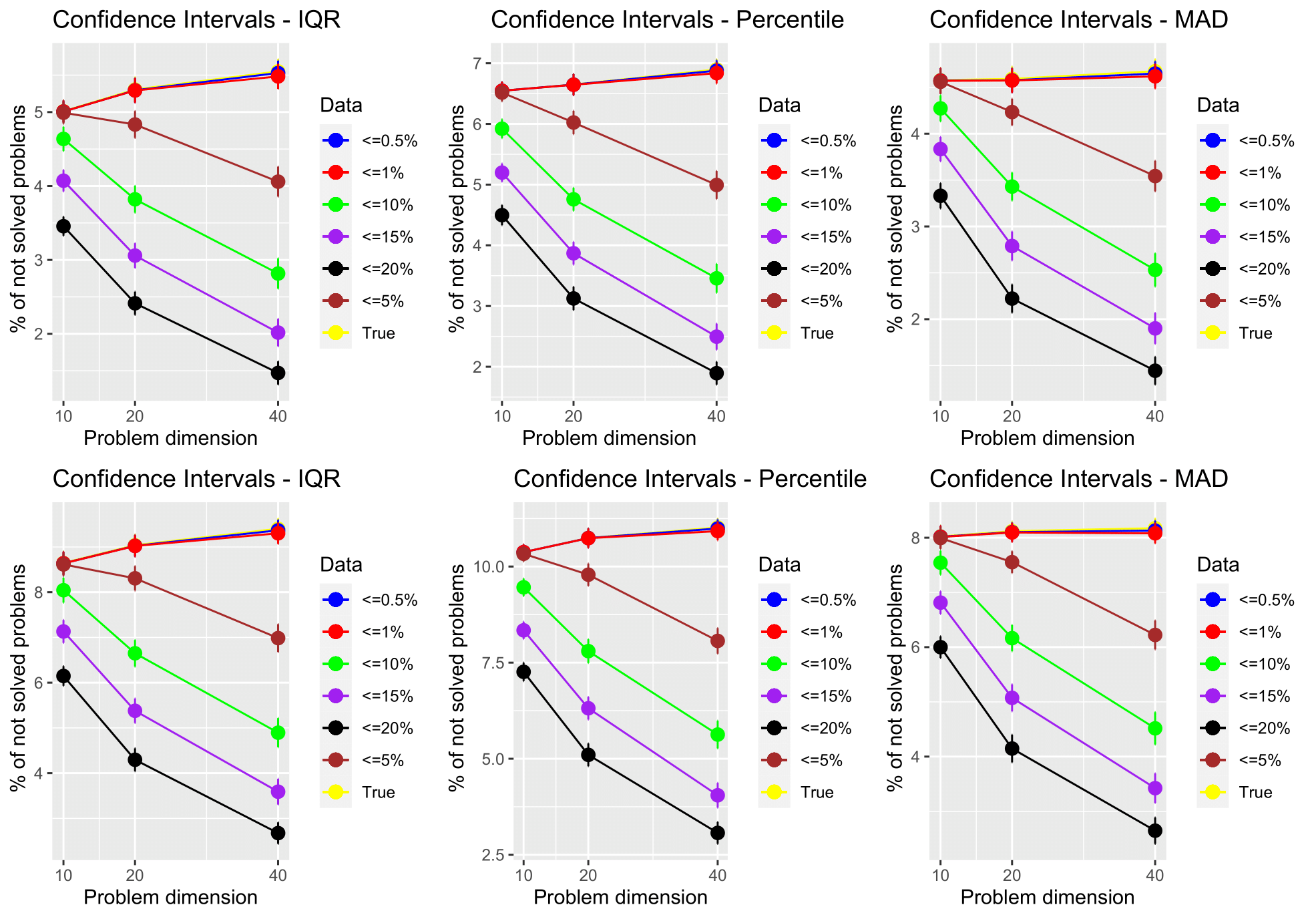}
      \caption{The error bars represent confidence intervals for the percentage of problems per dimension where the estimated runs do not produce accurate results, calculated across 104 DE configurations. The top row shows results for a skewness threshold of 0.05, the bottom row for 0.15, with columns for outlier detection techniques (\emph{IQR}, \emph{Percentile}, and \emph{MAD}).}
    \label{fig:error_bars}
\end{figure*}

To go into more detail, Figure~\ref{fig:error_bars} provides a detailed view with error bars representing the CIs for the percentage of triplets where the number of runs was not appropriately estimated, shown for each problem dimension. The percentages were first averaged across 10 sampling seeds for each DE configuration, and the error bars were then computed across 104 DE configurations. The top row shows results for a skewness threshold of 0.05, while the bottom row shows results for 0.15. Columns represent different outlier detection techniques: \emph{IQR"}, \emph{Percentile"}, and \emph{``MAD"}.

From the figure, we observe that, based on bootstrapping CIs or error ranges of $\leq 0.5\%$ and $\leq 1\%$ (i.e., post-hoc analysis), the percentage of triplets where the number of runs is not correctly estimated remains consistent across dimensions, ranging from 5\% to 7\% depending on the outlier detection method (in case of the skewness threshold set to 0.05). Larger error ranges ($\leq 5\%$, $\leq 10\%$, $\leq 15\%$, $\leq 20\%$) favor higher dimensions, as the percentage of incorrectly estimated triplets decreases with increased flexibility in error range. This occurs because, in higher dimensions, the collected results typically exhibit greater variance, leading to more widely dispersed solutions. When bootstrapping from such distributions, the mean value tends to fall within a specified range around the true mean. In contrast, in lower dimensions, solutions are more tightly clustered, and additional averaging can shift the bootstrap mean. Increasing the skewness threshold from 0.05 to 0.15 raises the percentages but maintains similar patterns, which also hold true for the omitted skewness thresholds.

\begin{figure*}[thb]
    \centering
    \includegraphics[scale=0.45]{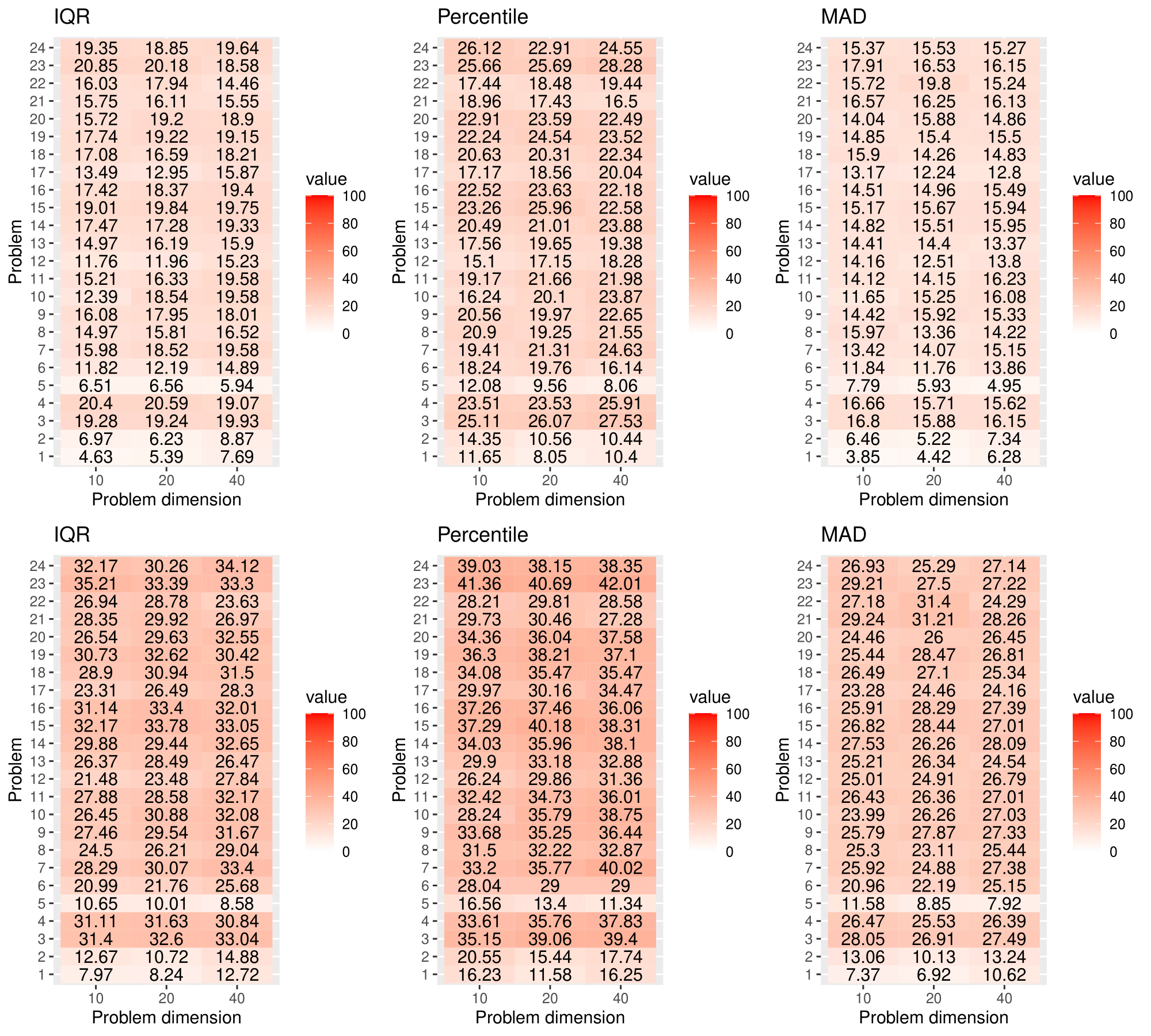}
    \caption{The percentage of triplets with inaccurate estimations across 104 DE configurations is presented by problem and dimension. The top row shows results for a skewness threshold of 0.05, the bottom row for 0.15, with columns representing outlier detection techniques (\emph{IQR}, \emph{Percentile}, and \emph{MAD}).}
    \label{fig:error_dim_percentage}
\end{figure*}

Next, for each problem and dimension, the percentage of triplets where the estimated number of runs fails to provide accurate results is calculated across 104 DE configurations. This percentage is determined by averaging the cases where the estimation is incorrect over 10 sampling seeds, separately for each DE configuration. For the COCO benchmark suite, the results are first averaged at the problem level (treating all instances of a problem as a single entity) before averaging across sampling seeds. The final results, averaged across all DE configurations for each problem and dimension, are shown in Figure~\ref{fig:error_dim_percentage}.

The top row corresponds to a skewness threshold of 0.05, and the bottom row to 0.15, with columns representing outlier detection techniques (\emph{IQR"}, \emph{Percentile"}, and \emph{``MAD"}). Each heatmap cell indicates the probability of failing to appropriately estimate the number of runs for a given problem and dimension. Lower percentages suggest a higher likelihood of accurate estimation.

The figure reveals that all problems across dimensions are represented in cases where the estimation fails, though functions like the sphere (problem 1) and linear slope (problem 5) in the COCO suite are easier to estimate accurately. By increasing the skewness threshold, the percentages increase which is an expected pattern. It is important to note that these results aggregate the varying behaviors of different DE configurations.

\noindent\textbf{Benchmarking the Nevergrad Algorithm Portfolio Across 20$d$ Problems:} 
\begin{figure}[tb]
    \centering
    \includegraphics[scale=0.42]{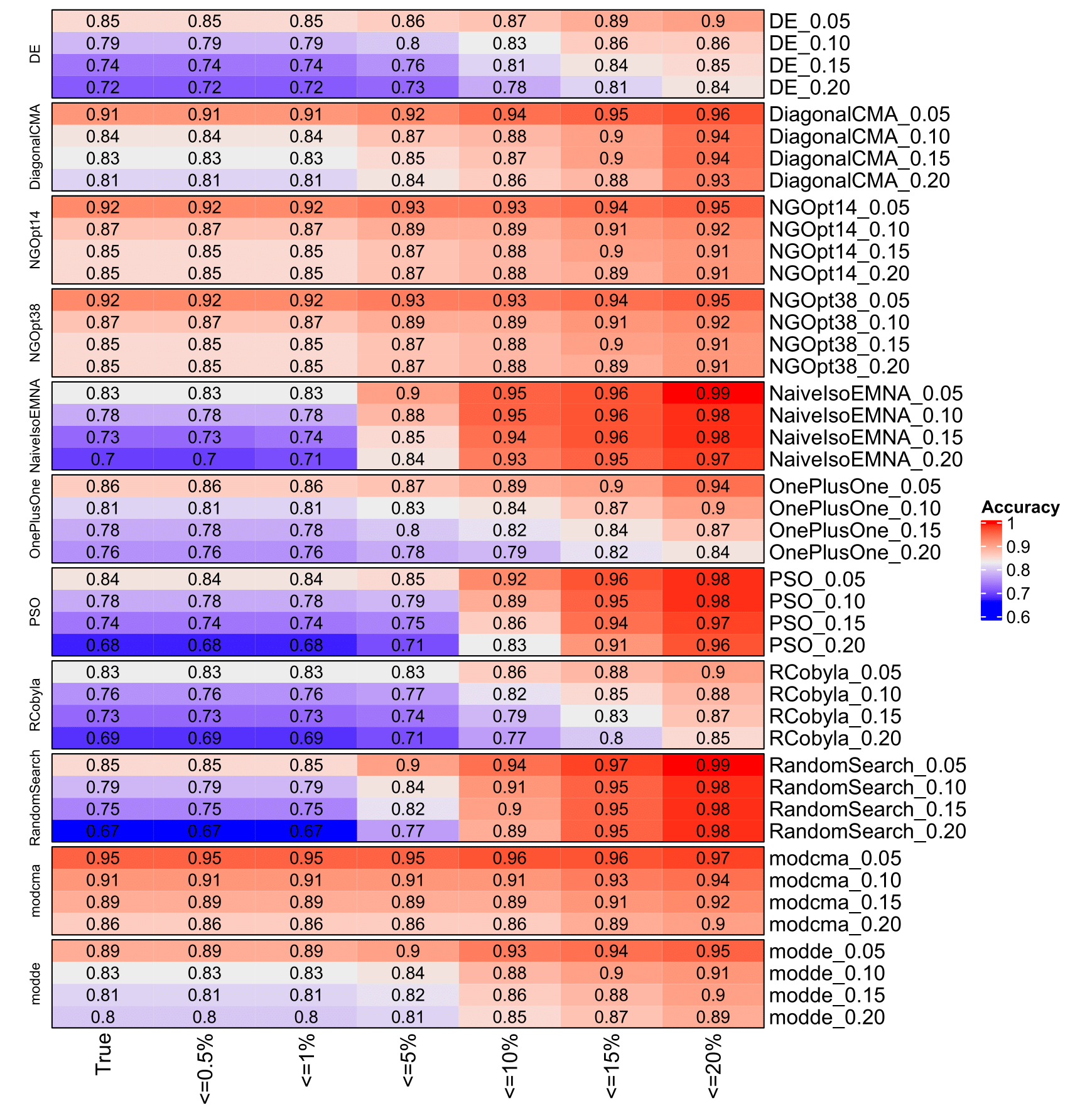}
    \caption{The heatmap shows the percentage of triplets where the proposed method correctly estimates the number of runs, using bootstrapped CIs and post-hoc evaluations. It is divided into 11 groups, each representing a different Nevergard algorithm, with results presented for skewness thresholds of 0.05, 0.10, 0.15, and 0.20.}
    \label{fig:Nevergrad_sensitivty}
    \vspace{-1em}
\end{figure}
Figure~\ref{fig:Nevergrad_sensitivty} shows the percentage of triplets for which the proposed method accurately estimates the required number of runs for the Nevergrad portfolio, using the \emph{``MAD"} outlier detection method (similar results are achived by the other methods). 
The results are grouped by 11 Nevergard algorithms and presented for skewness thresholds of 0.05, 0.10, 0.15, and 0.20. Focusing on the skewness threshold set to 0.05, the results indicate that the proposed method achieves accurate estimations, with percentages ranging from 82.58\% (RCobyla, 198/240 triplets correctly estimated) to around 95.16\% for modCMA, 91.12\% for diagonal CMA, 92.10\% for NGOpt14 and 92.08\% for NGOpt38. Post-hoc evaluation for RCobyla further improved accuracy to 85.54\% within a 10\% error to the ground truth mean. In all other algorithms, the post-hoc evaluation improves the results to approximately 90.00\% or even higher percentages in finding a solution within an error of 10\% to the mean of the ground truth sample. By increasing the skewness threshold, allowing more flexibility and deviations of the distribution symmetry, the estimation accuracy decreases, as expected. Nonetheless, accuracy remains high $\ge$ 80.00\%) when focusing on solutions within a 10\% error margin from the ground truth sample mean. These results are shown for the \emph{``MAD"} outlier detection method, with similar findings for other methods not presented due to the page limits.

\section{Green benchmarking}
\label{s:green}
Online estimation of the required number of runs for a given algorithm and problem instance supports energy-efficient experimentation, aligning with the principles of green benchmarking. This approach minimizes unnecessary computations, contributing to a more sustainable and environmentally friendly computing ecosystem. By enabling algorithm designers to focus on energy-efficient solutions, it fosters advancements toward a greener future in computing. In our study, we estimated the potential reduction in runs for each outlier detection technique across combinations of benchmark suites and algorithm portfolios. The results, summarized in Table~\ref{tab:com_analysis}, include metrics such as total runs (executed 50 times per triplet of problem, instance, and dimension), estimated runs, and saved runs, representing the difference between total and estimated runs. Saved runs highlight unnecessary computations that could be eliminated without affecting results, expressed as a percentage of total runs. To account for inaccuracies in estimation, we also report expected saved runs, calculated as the percentage of accurate estimations applied to saved runs. The percentage of accurate estimations is calculated as an aggregate across all algorithms in the analyzed portfolio. The top row of the table corresponds to the results obtained for the 104 DE configurations, while the bottom row of the table corresponds to the results obtained for the Nevergrad portfolio.

\begin{table*}[!htp]
\centering
\caption{Computational analysis.}
\label{tab:com_analysis}
\scriptsize
\begin{tabular}{lrrrrrrrrrrr}
\toprule
\begin{tabular}[c]{@{}l@{}}Algorithm \\ portfolio\end{tabular} &
\begin{tabular}[c]{@{}l@{}}Skewness \\ threshold\end{tabular} &
\begin{tabular}[c]{@{}l@{}}Outlier \\ technique\end{tabular} &
\begin{tabular}[c]{@{}l@{}}Total \\ runs\end{tabular} &
\begin{tabular}[c]{@{}l@{}}Estimated \\ runs\end{tabular} &
\begin{tabular}[c]{@{}l@{}}\% of \\ estimated \\ runs\end{tabular} &
\begin{tabular}[c]{@{}l@{}}Saved \\ runs\end{tabular} &
\begin{tabular}[c]{@{}l@{}}\% of \\ saved \\ runs\end{tabular} &
\begin{tabular}[c]{@{}l@{}}\% of \\ accurate \\ estimation\end{tabular} &
\begin{tabular}[c]{@{}l@{}}Expected \\ saved \\ runs\end{tabular} &
\begin{tabular}[c]{@{}l@{}}\% of \\ expected \\ saved \\ runs\end{tabular} \\
\midrule
\multirow{12}{*}{DE configurations} 
& 0.05 & IQR        & 5616000 & 2939131 & 52.33 & 2676869 & 47.67 & 84.13 & 2252087.64 & 40.10 \\
&      & Percentile & 5616000 & 2936222 & 52.28 & 2679778 & 47.72 & 79.91 & 2141341.12 & 38.13 \\
&      & MAD        & 5616000 & 3184780 & 56.71 & 2431220 & 43.29 & 86.16 & 2094654.47 & 37.30 \\
\cmidrule{2-11}
& 0.10 & IQR        & 5616000 & 2325231 & 41.40 & 3290769 & 58.60 & 77.21 & 2540911.97 & 45.24 \\
&      & Percentile & 5616000 & 2335983 & 41.60 & 3280017 & 58.40 & 72.36 & 2373337.02 & 42.26 \\
&      & MAD        & 5616000 & 2567520 & 45.72 & 3048480 & 54.28 & 79.72 & 2430286.14 & 43.27 \\
\cmidrule{2-11}
& 0.15 & IQR        & 5616000 & 1988204 & 35.40 & 3627796 & 64.60 & 72.91 & 2645093.50 & 47.10 \\
&      & Percentile & 5616000 & 2010232 & 35.79 & 3605768 & 64.21 & 67.86 & 2447049.06 & 43.57 \\
&      & MAD        & 5616000 & 2204761 & 39.26 & 3411239 & 60.74 & 75.68 & 2581738.78 & 45.97 \\
\cmidrule{2-11}
& 0.20 & IQR        & 5616000 & 1731759 & 30.84 & 3884241 & 69.16 & 69.50 & 2699723.86 & 48.07 \\
&      & Percentile & 5616000 & 1777237 & 31.65 & 3838763 & 68.35 & 64.63 & 2481128.69 & 44.18 \\
&      & MAD        & 5616000 & 1918048 & 34.15 & 3697952 & 65.85 & 72.37 & 2676172.17 & 47.65 \\
\bottomrule
\multirow{12}{*}{Nevergrad portfolio} 
& 0.05 & IQR        & 132000 & 59683 & 45.21 & 72317 & 54.79 & 85.14 & 61570.69 & 46.64 \\
&      & Percentile & 132000 & 59372 & 44.98 & 72628 & 55.02 & 81.05 & 58864.99  & 44.59 \\
&      & MAD        & 132000 & 67217 & 50.92 & 64783 & 49.08 & 87.81 & 56885.95 & 43.10 \\
\cmidrule{2-11}
& 0.10 & IQR        & 132000 & 46105 & 34.93 & 85895 & 65.07 & 78.59 & 67504.88 & 51.14 \\
&      & Percentile & 132000 & 46985 & 35.59 & 85015 & 64.41 & 74.26 & 63132.13  & 47.83 \\
&      & MAD        & 132000 & 53421 & 40.47 & 78579 & 59.53 & 82.19 & 64584.08 & 48.93 \\
\cmidrule{2-11}
& 0.15 & IQR        & 132000 & 39320 & 29.79 & 92680 & 70.21 & 75.70 & 70158.76   & 53.15 \\
&      & Percentile & 132000 & 39764 & 30.12 & 92236 & 69.88 & 70.85 & 65349.20  & 49.51 \\
&      & MAD        & 132000 & 46149 & 34.96 & 85851 & 65.04 & 79.15 & 67951.06 & 51.48 \\
\cmidrule{2-11}
& 0.20 & IQR        & 132000 & 34790 & 26.36 & 97210 & 73.64 & 73.11 & 71070.23  & 53.84 \\
&      & Percentile & 132000 & 35084 & 26.58 & 96916 & 73.42 & 67.57 & 65486.14 & 49.61 \\
&      & MAD        & 132000 & 40915 & 31.00 & 91085 & 69.00 & 76.28 & 69479.63  & 52.64 \\
\bottomrule
\end{tabular}
\end{table*}

From the table, we observe that increasing the skewness threshold reduces the number of required runs (i.e., saving a lot of runs) but at the cost of lower estimation accuracy. For instance, with a skewness threshold of 0.05 and the Nevergrad portfolio, the estimated number of required runs is highest across all outlier techniques. This indicates that achieving a closer approximation to a symmetric distribution demands more runs, resulting in greater accuracy. In this case, 43.10\% of the runs are expected to be unnecessary. Among the outlier techniques, all produce similar results; however, \emph{``MAD"} stands out as the most rigorous, estimating slightly more runs than the other methods while delivering the highest estimation accuracy. The same patterns are also visible for the DE configurations experiment.

This estimation is based exclusively on the evaluation of bootstrapping CIs. If post-hoc evaluation or identifying solutions within a predefined error range is considered, the percentage of estimated saved runs is likely to increase.

It is also important to highlight that this analysis provides a preliminary approximation. A more detailed investigation of individual problems and dimensions is needed for future research and deeper insights.

\section{Conclusion}
\label{s:conclusions}

Determining the optimal number of algorithm runs is critical in experimental setups, influencing both efficiency and result reliability. This paper presents a novel methodology for estimating the required runs in continuous single-objective stochastic optimization. Using probability theory and a robustness check to detect significant imbalances in the data distribution relative to the mean, the method dynamically adjusts the number of runs during execution. Evaluated across two algorithm portfolios and one benchmark suite with 5,748,000 runs, the methodology achieved an estimation accuracy of 82\%–95\%, reducing the number of runs by approximately 50\% without compromising optimization quality. This approach enhances efficiency and has the potential to support sustainable computing by minimizing energy consumption through online optimization.

A limitation of the proposed methodology is its occasional inaccuracy in estimating the required number of runs, with errors ranging from 5\% to 25\%, depending on the combination of an algorithm portfolio and a benchmark suite. Currently, these inaccuracies can only be identified after the evaluation is complete. However, our study has allowed us to annotate such cases. In future work, we plan to use these annotated cases to extract statistical features from the estimated samples, enriched with problem landscape characteristics~\cite{mersmann2011exploratory}, and algorithm hyperparameters. This enriched data will be used to train a supervised machine learning (ML) classifier capable of determining whether an estimation is accurate, along with a confidence probability. This advancement will address the limitation, enabling real-time implementation of the methodology in an online setting.
\begin{acks}
We acknowledge the support of the Slovenian Research and Innovation Agency through program grant P2-0098, and project grants No.J2-4460 and No. GC-0001. This work is also funded by the European Union under Grant Agreement 101187010 (HE ERA Chair AutoLearn-SI). We would like to thank Diederick Vermetten from the Leiden Institute of Advanced Computer Science (LIACS) in the Netherlands for providing us with the Nevergrad performance data. 
\end{acks}


\clearpage
\end{document}